\documentclass[10pt,twocolumn,letterpaper]{article}

\usepackage{cvpr}              

\usepackage{amsmath}
\usepackage{amssymb}
\usepackage{booktabs}
\usepackage{algorithm}
\usepackage{algorithmic}
\usepackage{verbatim}
\usepackage{mathrsfs}
\usepackage{marvosym}
\usepackage{multirow}
\usepackage{adjustbox}
\usepackage{graphicx}

\usepackage[pagebackref,breaklinks,colorlinks]{hyperref}

\usepackage[capitalize]{cleveref}
\crefname{section}{Sec.}{Secs.}
\Crefname{section}{Section}{Sections}
\Crefname{table}{Table}{Tables}
\crefname{table}{Tab.}{Tabs.}

\def\cvprPaperID{*****} 
\def\confName{CVPR}
\def\confYear{2023}

\begin{document}

\title{Boosting Cross-task Transferability of Adversarial Patches with Visual Relations}

\author{Tony Ma\textsuperscript{1}, Songze Li\textsuperscript{2,}\thanks{Intern student. \textsuperscript{\Letter} Corresponding author.}, Yisong Xiao\textsuperscript{2}, and Shunchang Liu\textsuperscript{2,3,}\textsuperscript{\Letter}\\
\textsuperscript{1}Shen Yuan Honors College, Beihang University\\
\textsuperscript{2}State Key Lab of Software Development Environmen, Beihang University\\
\textsuperscript{3}Zhongguancun Laboratory\\
{\tt\small \{tonyyyma\}@gmail.com}, 
 {\tt\small \{20373043, xiaoyisong, liusc\}@buaa.edu.cn}}

\maketitle

\begin{abstract}
The transferability of adversarial examples is a crucial aspect of evaluating the robustness of deep learning systems, particularly in black-box scenarios. Although several methods have been proposed to enhance cross-model transferability, little attention has been paid to the transferability of adversarial examples across different tasks. This issue has become increasingly relevant with the emergence of foundational multi-task AI systems such as Visual ChatGPT, rendering the utility of adversarial samples generated by a single task relatively limited. Furthermore, these systems often entail inferential functions beyond mere recognition-like tasks. To address this gap, we propose a novel \textbf{V}isual \textbf{R}elation-based cross-task \textbf{A}dversarial \textbf{P}atch generation method called \textbf{VRAP}, which aims to evaluate the robustness of various visual tasks, especially those involving visual reasoning, such as Visual Question Answering and Image Captioning. VRAP employs scene graphs to combine object recognition-based deception with predicate-based relations elimination, thereby disrupting the visual reasoning information shared among inferential tasks. Our extensive experiments demonstrate that VRAP significantly surpasses previous methods in terms of black-box transferability across diverse visual reasoning tasks.
\end{abstract}

\section{Introduction}
\label{sec:intro}
Adversarial examples are input samples that have been intentionally perturbed to deceive deep learning models \cite{goodfellow2014explaining,liu2019perceptual}. They have become a vital tool in evaluating the robustness of machine learning systems \cite{tang2021robustart,zhang2021interpreting}. An essential consideration when assessing the effectiveness of adversarial examples is their transferability, which refers to their ability to generalize across different models. Numerous studies have demonstrated that strong black-box transferability of adversarial samples is critical for assessing the robustness of deep learning models in real-world settings \cite{liu2022harnessing, wang2021dual, wang2022defensive}.

There has been considerable research aimed at improving the transferability of adversarial examples across models \cite{dong2018boosting, wang2022generating,liu2020bias}, however, very little work has focused on their transferability across tasks. Recently, some foundational multi-task visual systems such as Visual ChatGPT \cite{visualcahtgpt} and GPT-4 \cite{gpt4} have emerged. These systems no longer rely on a single machine-learning model to handle specific tasks, but rather, can solve problems in different scenarios,  leading to a reduction in the effectiveness of adversarial examples depending on a single task. Moreover, with the continued development of artificial intelligence, reasoning tasks are expected to become the mainstream tasks of these foundational models. Visual reasoning tasks encompass a wide range of task paradigms, requiring neural network models not only to recognize objects in images but also to understand the relationships between objects and use this information to complete downstream tasks. Thus, visual reasoning tasks provide more prior information than simple visual recognition tasks, and the combination of visual representation and reasoning prior is a promising approach for robust multi-task integration \cite{yang2022logicdef}. However, existing cross-task attack methods only consider attacks on basic visual recognition tasks \cite{zhang2022enhance,lu2020enhancing}, and do not take into account the possible visual reasoning information, making them less effective when attacking the aforementioned multi-task models.

To address the problem, it is important to develop a method that is able to generate adversarial examples with strong cross-task transferability, particularly in the domain of cross-reasoning-task transferability. To achieve this, we propose a novel approach called VRAP, which is a visual relation-based cross-task adversarial patch generation method. The central idea behind VRAP is to enable the threatening model to learn the potential relation information present within the image. In doing so, we seek to disrupt the visual reasoning information shared across various inferential tasks, resulting in a positive effect on enhancing the cross-task transferability. VRAP utilizes scene graphs \cite{xu2020survey}, a data structure that represents the objects in a visual scene and their relationships with one another, to combine object recognition-based deception with predicate-based relations elimination. Specifically, the approach leverages object detection to identify objects within the image and then creates a scene graph that represents the relations between these objects. This graph is then used to generate an adversarial patch that is designed to disrupt the relations between the objects. Based on this, we can generate adversarial patches with strong cross-task transferability. Our extensive experiments demonstrate that VRAP outperforms previous methods in terms of black-box transferability across diverse visual reasoning tasks.

\section{Preliminary and Method}

Recent research suggests that the transferability of adversarial attacks is largely influenced by task-invariant characteristics \cite{zhang2022enhance, wang2021universal}. In this paper, we aim to identify task-shared characteristics that significantly impact model performance. Specifically, we focus on visual reasoning tasks and seek to enable the attacker to learn the potential relation information present within the image. To achieve this, we propose a novel approach for generating an adversarial perturbation $\delta$ constrained to a localized patch, which can fool visual reasoning models into making incorrect predictions.

Specifically, given a clean image $x$, an additive adversarial patch perturbation $\delta$ $\in \mathbb{R}^z$, and a location mask $M$ $\in$ \{0,1\}$^n$, we can generate an adversarial example $x_{adv}$ as following,
\begin{equation} 
    x_{adv} = (1- M)\odot x + M \odot \delta, \enspace s.t. \enspace f(x_{adv}) \neq y,
\label{eqn:xadv}
\end{equation}
where $\odot$ is the element-wise multiplication, $f$ is a visual reasoning model, and $y$ is corresponding ground truth.

To obtain a better understanding of the relationships shared by visual reasoning tasks, we introduce a scene graph generation model $f_{G}$. Scene graphs provide a structured representation of the visual content by modeling the relationships between objects and their attributes using predicates, such as a car ``on" a bridge or a car ``has" wheels. Specifically, the scene graph generation model consists of two parts: an object detection module and a predicate classification module. The object detection module detects and localizes objects within the image, and the predicate classification module assigns predicates to the relationships between the detected objects. By feeding the adversarial example into the model $f_{G}$, we can obtain two probabilities, ${P}_C \in \mathbb{R}^{n \times m_c}$ and $P_R \in \mathbb{R}^{n \times {n-1} \times m_r}$, representing the object classification probability and the predicate classification probability, respectively. $n$ is the number of predicted targets, $m_c$ is the number of target labels, and $m_r$ is the number of predicate labels.

To effectively break the relationships between objects in an image, we propose the use of a relationship elimination loss $\mathcal L_r$. This loss term enables the adversarial patch to capture the relationship features between objects by reducing the probability of predicted relationships between all pairs of objects. Specifically, the relationship elimination loss can be formulated as:
\begin{equation}
    \mathcal L_r = \sum_{\mu, \nu} \max {P_R({x}_{adv} | \mu, \nu)},
\label{lr-loss}
\end{equation}
where $\mu$,$\nu$ represent object pairs. The loss is computed by adding up the scores of the most likely type of each relation between object pairs. This aggregated score represents the total relation-related score of the entire relation graph. By minimizing this loss, we encourage the patch to disrupt the relationships between objects, thereby increasing the chances of the reasoning information being able to successfully evade recognition by visual reasoning models.

In addition to breaking the relationships between objects in an image, we also explore the use of a target detection deception loss $\mathcal{L}_{d}$ to mislead object detection and classification models. This loss term is designed to alter the prediction of the model from the perspective of the misleading model detection labels. Specifically, it can be formulated as:
\begin{equation}
    \mathcal L_d = \sum_{i} \max {P_C({x}_{adv} | y_i^{{f_G}} = y_i^{GT})},
\end{equation}
where $y_i^{f_G}$ represents the predicted label of the model and $y_i^{GT}$ represents the real label of the target object $i$.
By doing so, we aim to deceive the model into misidentifying the target objects in the image.

The generation of adversarial perturbations is achieved by utilizing a combination of two losses, \ie, 
\begin{equation}
    \arg\min_{\delta}{\mathcal L_{d} + \lambda \mathcal L_{r}},
\end{equation}
where $\lambda$ controls the contributions of each term. The approach employed utilizes a gradient-based iterative algorithm to optimize the adversarial patches. In each iteration, an initial adversarial patch is generated at a random position. Subsequently, a forward pass is conducted to obtain the bounding box feature and the corresponding target detection deception loss $\mathcal{L}_{d}$. The visual relation graph is then computed, and the relationship elimination loss $\mathcal{L}_{r}$ is obtained. Finally, the adversarial patch is updated using the back-propagation algorithm, enabling it to attack the shared relation information of visual reasoning models. Thus, the resulting perturbation exhibits a high degree of transferability across visual reasoning tasks. 

\section{Experiment}

\begin{table*}[!ht]
\setlength{\belowcaptionskip}{-0.2cm}
\setlength{\abovecaptionskip}{-0.3cm}
\caption{Attacking Results on the SGG task. Lower R@K and mR@K means stronger attack.}
\begin{center}
\setlength{\tabcolsep}{18.5pt}
\scriptsize
\begin{tabular}{cccccccc}
\toprule
{\textbf{Subtasks}}&{\textbf{Method}}&{\textbf{R@20}}&{\textbf{R@50}}&{\textbf{R@100}}&{\textbf{mR@20}}&{\textbf{mR@50}}&{\textbf{mR@100}}\\
\midrule
\multirow{3}{*}{PredCls}
&{Raw}
&{59.64} 
&{66.12}
&{67.97} 
&{11.44}
&{14.59}
&{15.84} \\
\multirow{3}{*}{}
&{DR}
&{59.42} 
&{65.98}
&{67.86} 
&{11.22}
&{14.26}
&{15.47} \\
\multirow{3}{*}{}
&{\textbf{VRAP}}
&\textbf{59.30}
&\textbf{65.94} 
&\textbf{67.83} 
&\textbf{11.18} 
&\textbf{14.25} 
&\textbf{15.44} \\
\midrule
\multirow{3}{*}{SGCls}
&{Raw}
&{36.00} 
&{39.24}
&{40.05} 
&{6.49}
&{8.02}
&{8.50} \\
\multirow{3}{*}{}
&{DR}
&{34.08} 
&{37.07}
&{37.83} 
&{6.15}
&{7.50}
&{7.97} \\
\multirow{3}{*}{}
&{\textbf{VRAP}}
&\textbf{33.63}
&\textbf{36.61} 
&\textbf{37.35} 
&\textbf{6.02} 
&\textbf{7.41} 
&\textbf{7.85} \\
\midrule
\multirow{3}{*}{SGDet}
&{Raw}
&{25.40} 
&{32.45}
&{37.24} 
&{4.37}
&{5.80}
&{7.06} \\
\multirow{3}{*}{}
&{DR}
&{23.75} 
&{30.02}
&{34.39} 
&{4.11}
&{5.39}
&{6.53} \\
\multirow{3}{*}{}
&{\textbf{VRAP}}
&\textbf{21.96}
&\textbf{28.61} 
&\textbf{32.98} 
&\textbf{3.69} 
&\textbf{5.02} 
&\textbf{6.09} \\
\bottomrule
\end{tabular}
\end{center}
\vskip -2ex
\label{tab:SGG}
\end{table*}

\begin{table*}[!ht]
\setlength{\belowcaptionskip}{-0.2cm}
\setlength{\abovecaptionskip}{-0.3cm}
\caption{Attacking Results on the IC task. Lower values of the metrics means stronger attack}
\begin{center}
\setlength{\tabcolsep}{15pt}
\scriptsize
\begin{tabular}{ccccccccc}
\toprule
{\textbf{Method}}&{\textbf{Bleu\_1}}&
{\textbf{Bleu\_2}}&
{\textbf{Bleu\_3}}&
{\textbf{Bleu\_4}}&{\textbf{METEOR}}&{\textbf{ROUGH\_L}}&{\textbf{CIDEr}}&{\textbf{SPICE}}\\
\midrule
{Raw}
&{83.75} 
&{69.32}
&{54.97}
&{42.60}
&{31.33} 
&{61.53} 
&{146.49} 
&{25.48}\\
{DR}
&{81.53}
&{66.71}
&{52.41}
&{40.22}
&{30.45} 
&{60.03} 
&{\textbf{138.16}}
&{24.30}\\
{\textbf{VRAP}}
&{\textbf{81.19}}
&{\textbf{66.35}}
&{\textbf{52.07}}
&{\textbf{39.94}}
&{\textbf{30.27}} 
&{\textbf{59.80}} 
&{138.62} 
&{\textbf{24.20}}\\
\bottomrule
\end{tabular}
\end{center}
\vskip -2ex
\label{tab:OFA}
\end{table*}


\textbf{Datasets.}
We use the Visual Genome dataset \cite{krishna2017visual}, a giant visual relation dataset with 108,077 images and 2.3 million relationships, for patch generation. Besides, we also use VQAv2 dataset \cite{goyal2017making} and COCO \cite{lin2014microsoft} dataset for evaluation. 

\textbf{Target tasks and models.}
We conduct attacks towards three typical visual reasoning tasks: Scene Graph Generation (SGG), Vision Questions
Answering (VQA), and Image Captioning (IC) to verify our method. SGG consists of 3 different subtasks, including Predicate Classification (PredCls), Scene Graph Classification (SGCls) and Scene Graph Detection (SGDet). The inputs to these subtasks are different. PredCls uses ground-truth bounding boxes and object labels as inputs, SGCls uses only ground-truth bounding boxes, and SGDet does not require any ground-truth information. For SGG, we conduct attack on classical causal Neural-MOTIFS model \cite{tang2020unbiased}. For VQA and IC, we take the OFA-Base\cite{wang2022ofa} as the target model. It is a unified sequence-to-sequence pretrained model that unifies modalities and tasks, which achieves the state-of-the-art performance at the COCO Leaderboard for image captioning task.

\textbf{Evaluation metrics.}
 For SGG, we use Recall@K (R@K) and Mean Recall@K (mR@K) as our metrics following \cite{tang2020unbiased}. The values of K are taken as 20, 50 and 100, respectively. For IC, we use the metrics following\cite{wang2022ofa}.

\textbf{Baselines.} To the best of our knowledge, we are the first work to study the transferability of adversarial examples across visual reasoning tasks. In this paper, we compare only with the state-of-the-art attack approach, Dispersion Reduction (DR) \cite{lu2020enhancing}, aiming to transfer across recognition tasks. We will consider more baselines in future.

\textbf{Implementation details.}
For patch generation, we only use the validation set of Visual Genome which contains 5000 images for the training stage due to the time consumption.  We randomly initialize a $80\times80$ square adversarial patch and conduct training with batch size 1 by $T = 5$ iterations every epoch with an attack step size $\alpha$ of 0.04, and a maximum of 2 epochs. The position and orientation of the patch are randomly chosen, which makes our adversarial patches able to universally attack all scenes. For our method, we set the relation loss weight $\lambda$ as 0.01. For DR, we choose to attack the dispersion of feature after the third block of backbone. All of our codes are implemented in PyTorch. We conduct all experiments on a NVIDIA GTX2080Ti GPU with 10GB Memory.

\subsection{Attacking Results on SGG}
First we conduct experiments for scene graph generation task. We generate adversarial patches using the SGCls model and further perform attacks on each subtask accordingly. As shown in Table \ref{tab:SGG}, in contrast to DR, our method achieves lower R@K and mR@K on both white-box and black-box settings, which means stronger attacking ability.

\subsection{Attacking Results on VQA and IC}
\begin{figure*}[!ht]
\setlength{\belowcaptionskip}{-0.05cm}
\setlength{\abovecaptionskip}{-0.1cm}
\begin{center}
\includegraphics[width=0.95\linewidth]{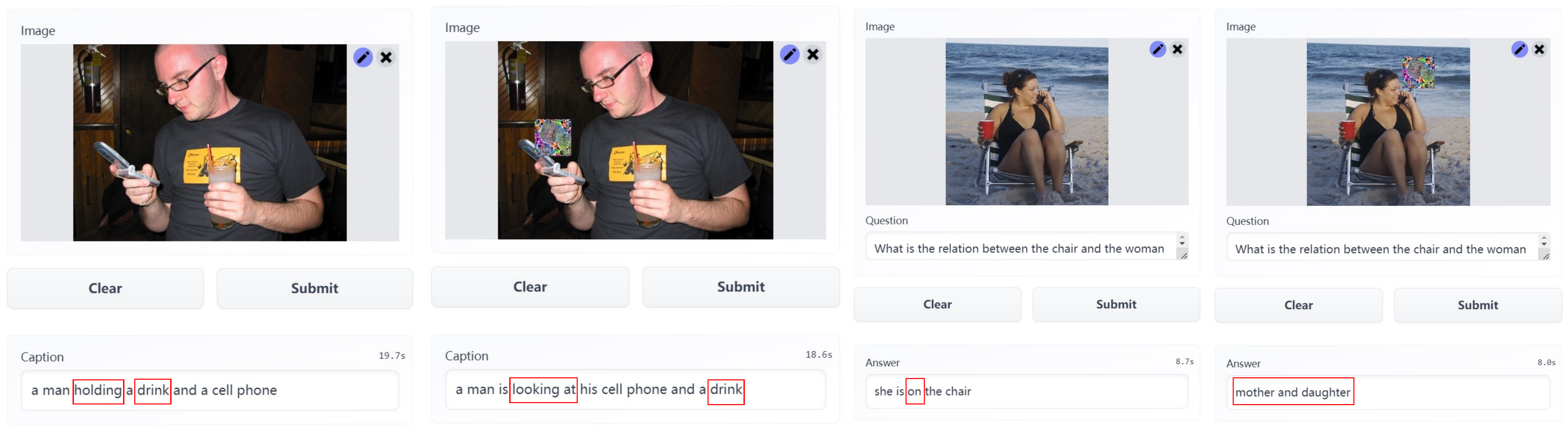}
\end{center} 
\vskip -2ex
\caption{Our patch can successfully mislead the model relation prediction in image caption and visual question answering.} 
\label{fig:1}
\vskip -2ex
\end{figure*}

For VQA and IC, we directly pasted the patches generated from the SGG task on the test images for black-box evaluation. Table \ref{tab:OFA} shows that our method almost achieves lower accuracy for the image caption task, leading to a significantly stronger black-box attacking transferability. We also visualize the results of IC and VQA through the Hugging Face spaces \footnote{https://huggingface.co/OFA-Sys}. As shown in Figure \ref{fig:1}, it can be seen that our adversarial patch successfully misled the model's judgment towards the relationship.

\section{Conclusion}

In this paper, we present a novel cross-task adversarial patch  generation method, named VRAP, to evaluate the robustness of various visual reasoning tasks. VRAP leverages scene graphs to disrupt the shared information of those tasks by combining object recognition-based deception with predicate-based relations elimination. The experiments conducted show that VRAP outperforms previous methods in terms of black-box transferability. Future research can explore the potential of VRAP in other visual tasks and its application in real-world scenarios.

{\small
\bibliographystyle{ieee_fullname}

}

\end{document}